\documentclass[10pt,twocolumn,letterpaper]{article}

\usepackage{iccv,times,graphicx,amsmath,amssymb,booktabs,tabulary,multirow,overpic,xcolor}
\usepackage[caption=false]{subfig}
\definecolor{citecolor}{RGB}{34,139,34}
\usepackage[pagebackref=true,breaklinks=true,letterpaper=true,colorlinks,
  citecolor=citecolor,bookmarks=false]{hyperref}
\usepackage[british,english,american]{babel}
\def\iccvPaperID{1902}\iccvfinalcopy

\newcommand{\bd}[1]{\textbf{#1}}
\newcommand{\app}{\raise.17ex\hbox{$\scriptstyle\sim$}}
\newcommand{\symb}[1]{{\small\texttt{#1}}\xspace}
\newcommand{\mrtwo}[1]{\multirow{2}{*}{#1}}
\def\x{\times}
\def\pt{p_\textrm{t}}
\def\at{\alpha_\textrm{t}}
\def\xt{x_\textrm{t}}
\def\CE{\textrm{CE}}
\def\FL{\textrm{FL}}
\def\FQ{\textrm{FL}^*}
\newcommand{\eqnnm}[2]{\begin{equation}\label{eq:#1}#2\end{equation}\ignorespaces}
\newcolumntype{x}[1]{>{\centering\arraybackslash}p{#1pt}}
\newlength\savewidth\newcommand\shline{\noalign{\global\savewidth\arrayrulewidth
  \global\arrayrulewidth 1pt}\hline\noalign{\global\arrayrulewidth\savewidth}}
\newcommand{\tablestyle}[2]{\setlength{\tabcolsep}{#1}\renewcommand{\arraystretch}{#2}\centering\footnotesize}
\makeatletter\renewcommand\paragraph{\@startsection{paragraph}{4}{\z@}
  {.5em \@plus1ex \@minus.2ex}{-.5em}{\normalfont\normalsize\bfseries}}\makeatother

\begin{document}

\title{Focal Loss for Dense Object Detection\vspace{-5mm}}
\author{
 Tsung-Yi Lin \quad  Priya Goyal \quad Ross Girshick \quad Kaiming He \quad Piotr Doll\'ar \vspace{1mm}\\
 Facebook AI Research (FAIR)
}
\maketitle

\begin{figure}[t]
\centering
\begin{overpic}[width=.99\linewidth]{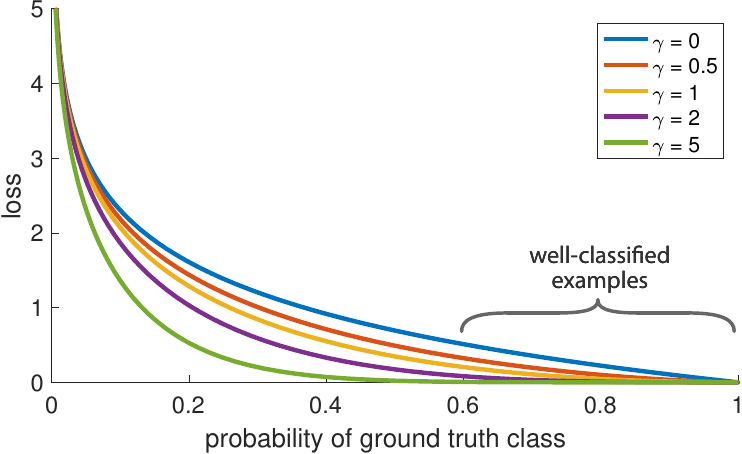}
 \put(18, 54){\small $\CE(\pt) = - \log (\pt)$}
 \put(19, 48){\small $\FL(\pt) = -(1 - \pt)^\gamma \log (\pt)$}
\end{overpic}
\caption{We propose a novel loss we term the \emph{Focal Loss} that adds a factor $(1 - \pt)^\gamma$ to the standard cross entropy criterion. Setting $\gamma>0$ reduces the relative loss for well-classified examples ($\pt>.5$), putting more focus on hard, misclassified examples. As our experiments will demonstrate, the proposed focal loss enables training highly accurate dense object detectors in the presence of vast numbers of easy background examples.}
\label{fig:loss}
\end{figure}

\begin{abstract}\vspace{-1mm}
The highest accuracy object detectors to date are based on a two-stage approach popularized by R-CNN, where a classifier is applied to a \emph{sparse} set of candidate object locations. In contrast, one-stage detectors that are applied over a regular, \emph{dense} sampling of possible object locations have the potential to be faster and simpler, but have trailed the accuracy of two-stage detectors thus far. In this paper, we investigate why this is the case. We discover that the extreme foreground-background class imbalance encountered during training of dense detectors is the central cause. We propose to address this class imbalance by reshaping the standard cross entropy loss such that it down-weights the loss assigned to well-classified examples. Our novel \emph{Focal Loss} focuses training on a sparse set of hard examples and prevents the vast number of easy negatives from overwhelming the detector during training. To evaluate the effectiveness of our loss, we design and train a simple dense detector we call RetinaNet. Our results show that when trained with the focal loss, RetinaNet is able to match the speed of previous one-stage detectors while surpassing the accuracy of all existing state-of-the-art two-stage detectors. Code is at: {\footnotesize\url{https://github.com/facebookresearch/Detectron}}.
\end{abstract}\vspace{-5mm}

\begin{figure}[t]
\includegraphics[width=1\linewidth]{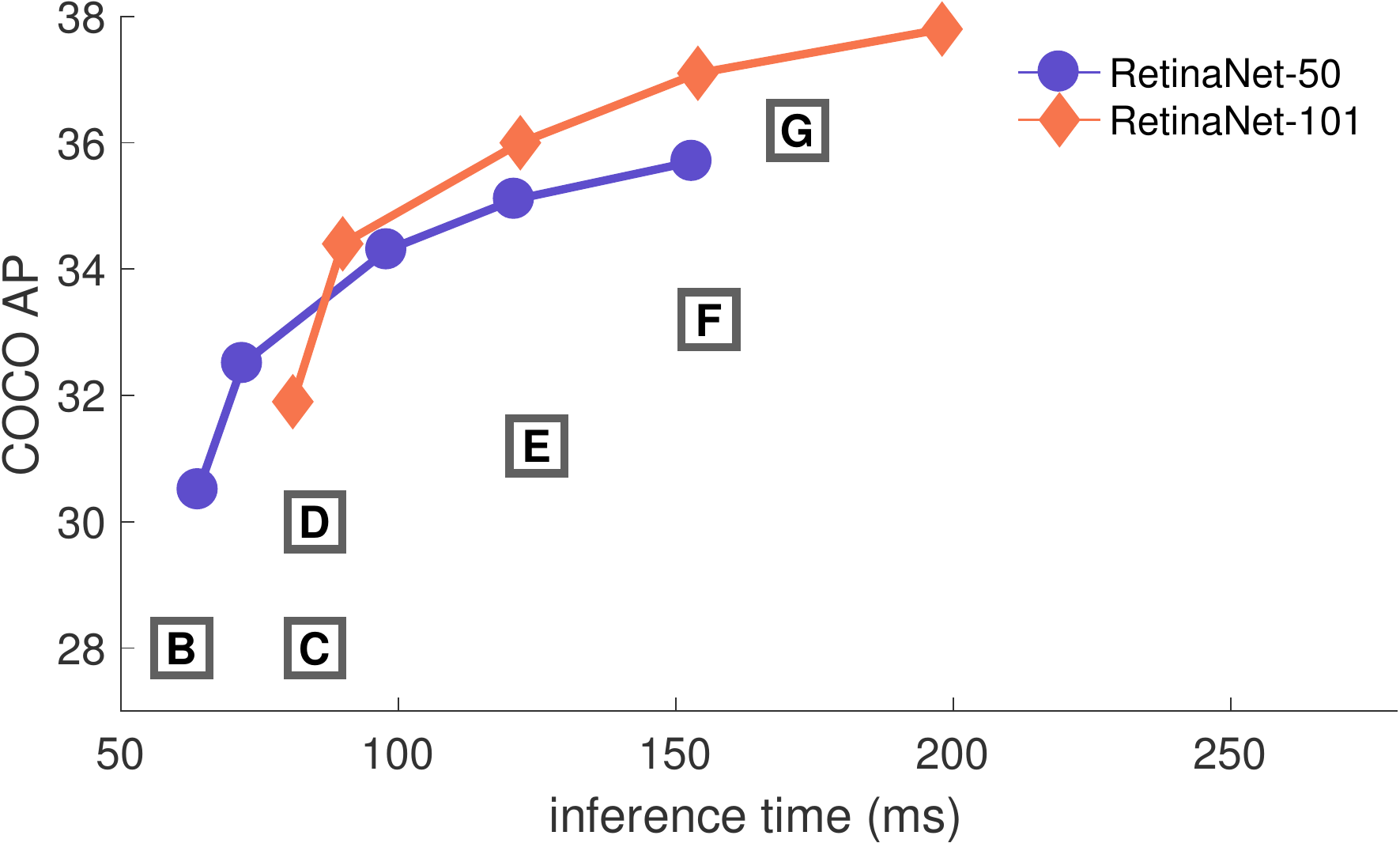}
\hspace{-35mm}\resizebox{.37\columnwidth}{!}{\tablestyle{2pt}{1}
\begin{tabular}[b]{l|cc}
 & AP & time \\
 \shline
 \bd{[A]} YOLOv2$^\dagger$ \cite{Redmon2017} & 21.6 & 25 \\
 \bd{[B]} SSD321   \cite{Liu2016}    & 28.0 &  61 \\
 \bd{[C]} DSSD321  \cite{Fu2016}     & 28.0 &  85 \\
 \bd{[D]} R-FCN$^\ddagger$ \cite{Dai2016b}   & 29.9 &  85 \\
 \bd{[E]} SSD513   \cite{Liu2016}    & 31.2 & 125 \\
 \bd{[F]} DSSD513  \cite{Fu2016}     & 33.2 & 156 \\
 \bd{[G]} FPN FRCN \cite{Lin2016}    & 36.2 & 172 \\
 \hline
 \bd{RetinaNet-50-500}  & 32.5 & 73  \\
 \bd{RetinaNet-101-500} & 34.4 & 90  \\
 \bd{RetinaNet-101-800} & 37.8 & 198 \\
 \multicolumn{3}{l}{~~$^\dagger$Not plotted ~~$^\ddagger$Extrapolated time\vspace{13mm} } \\
\end{tabular}}
\caption{Speed (ms) versus accuracy (AP) on COCO \texttt{test-dev}. Enabled by the focal loss, our simple one-stage \emph{RetinaNet} detector outperforms all previous one-stage and two-stage detectors, including the best reported Faster R-CNN \cite{Ren2015a} system from \cite{Lin2016}. We show variants of RetinaNet with ResNet-50-FPN (blue circles) and ResNet-101-FPN (orange diamonds) at five scales (400-800 pixels). Ignoring the low-accuracy regime (AP$<$25), RetinaNet forms an upper envelope of all current detectors, and an improved variant (not shown) achieves 40.8 AP. Details are given in \S\ref{sec:exps}.}
\label{fig:speed}
\end{figure}

\section{Introduction}

Current state-of-the-art object detectors are based on a two-stage, proposal-driven mechanism. As popularized in the R-CNN framework \cite{Girshick2014}, the first stage generates a \emph{sparse} set of candidate object locations and the second stage classifies each candidate location as one of the foreground classes or as background using a convolutional neural network. Through a sequence of advances \cite{Girshick2015a, Ren2015a, Lin2016, He2017}, this two-stage framework consistently achieves top accuracy on the challenging COCO benchmark \cite{Lin2014}.

Despite the success of two-stage detectors, a natural question to ask is: could a simple one-stage detector achieve similar accuracy? One stage detectors are applied over a regular, \emph{dense} sampling of object locations, scales, and aspect ratios. Recent work on one-stage detectors, such as YOLO \cite{Redmon2016, Redmon2017} and SSD \cite{Liu2016, Fu2016}, demonstrates promising results, yielding faster detectors with accuracy within 10-40\% relative to state-of-the-art two-stage methods.

This paper pushes the envelop further: we present a one-stage object detector that, for the first time, matches the state-of-the-art COCO AP of more complex two-stage detectors, such as the Feature Pyramid Network (FPN) \cite{Lin2016} or Mask R-CNN \cite{He2017} variants of Faster R-CNN \cite{Ren2015a}. To achieve this result, we identify class imbalance during training as the main obstacle impeding one-stage detector from achieving state-of-the-art accuracy and propose a new loss function that eliminates this barrier.

Class imbalance is addressed in R-CNN-like detectors by a two-stage cascade and sampling heuristics. The proposal stage (\eg, Selective Search \cite{Uijlings2013}, EdgeBoxes \cite{Zitnick2014}, DeepMask \cite{Pinheiro2015, Pinheiro2016}, RPN \cite{Ren2015a}) rapidly narrows down the number of candidate object locations to a small number (\eg, 1-2k), filtering out most background samples. In the second classification stage, sampling heuristics, such as a fixed foreground-to-background ratio (1:3), or online hard example mining (OHEM) \cite{Shrivastava2016}, are performed to maintain a manageable balance between foreground and background.

In contrast, a one-stage detector must process a much larger set of candidate object locations regularly sampled across an image. In practice this often amounts to enumerating \app 100k locations that densely cover spatial positions, scales, and aspect ratios. While similar sampling heuristics may also be applied, they are inefficient as the training procedure is still dominated by easily classified background examples. This inefficiency is a classic problem in object detection that is typically addressed via techniques such as bootstrapping \cite{Sung1994, Rowley95} or hard example mining \cite{Viola2001, Felzenszwalb2010a, Shrivastava2016}.

In this paper, we propose a new loss function that acts as a more effective alternative to previous approaches for dealing with class imbalance. The loss function is a dynamically scaled cross entropy loss, where the scaling factor decays to zero as confidence in the correct class increases, see Figure~\ref{fig:loss}. Intuitively, this scaling factor can automatically down-weight the contribution of easy examples during training and rapidly focus the model on hard examples. Experiments show that our proposed \emph{Focal Loss} enables us to train a high-accuracy, one-stage detector that significantly outperforms the alternatives of training with the sampling heuristics or hard example mining, the previous state-of-the-art techniques for training one-stage detectors. Finally, we note that the exact form of the focal loss is not crucial, and we show other instantiations can achieve similar results.

To demonstrate the effectiveness of the proposed focal loss, we design a simple one-stage object detector called \emph{RetinaNet}, named for its dense sampling of object locations in an input image. Its design features an efficient in-network feature pyramid and use of anchor boxes. It draws on a variety of recent ideas from \cite{Liu2016, Erhan2014, Ren2015a, Lin2016}. RetinaNet is efficient and accurate; our best model, based on a ResNet-101-FPN backbone, achieves a COCO \symb{test-dev} AP of 39.1 while running at 5 fps, surpassing the previously best published single-model results from both one and two-stage detectors, see Figure~\ref{fig:speed}.

\section{Related Work}

\paragraph{Classic Object Detectors:} The sliding-window paradigm, in which a classifier is applied on a dense image grid, has a long and rich history. One of the earliest successes is the classic work of LeCun \etal who applied convolutional neural networks to handwritten digit recognition \cite{LeCun1989, Lecun94}. Viola and Jones \cite{Viola2001} used boosted object detectors for face detection, leading to widespread adoption of such models. The introduction of HOG \cite{Dalal2005} and integral channel features \cite{Dollar2009} gave rise to effective methods for pedestrian detection. DPMs \cite{Felzenszwalb2010a} helped extend dense detectors to more general object categories and had top results on PASCAL \cite{Everingham2010} for many years. While the sliding-window approach was the leading detection paradigm in classic computer vision, with the resurgence of deep learning \cite{Krizhevsky2012}, two-stage detectors, described next, quickly came to dominate object detection.

\paragraph{Two-stage Detectors:} The dominant paradigm in modern object detection is based on a two-stage approach. As pioneered in the Selective Search work \cite{Uijlings2013}, the first stage generates a sparse set of candidate proposals that should contain all objects while filtering out the majority of negative locations, and the second stage classifies the proposals into foreground classes / background. R-CNN \cite{Girshick2014} upgraded the second-stage classifier to a convolutional network yielding large gains in accuracy and ushering in the modern era of object detection. R-CNN was improved over the years, both in terms of speed \cite{He2014, Girshick2015a} and by using learned object proposals \cite{Erhan2014, Pinheiro2015, Ren2015a}. Region Proposal Networks (RPN) integrated proposal generation with the second-stage classifier into a single convolution network, forming the Faster R-CNN framework \cite{Ren2015a}. Numerous extensions to this framework have been proposed, \eg \cite{Lin2016, Shrivastava2016, Shrivastava2016a, He2016, He2017}.

\paragraph{One-stage Detectors:} OverFeat \cite{Sermanet2014} was one of the first modern one-stage object detector based on deep networks. More recently SSD \cite{Liu2016, Fu2016} and YOLO \cite{Redmon2016,Redmon2017} have renewed interest in one-stage methods. These detectors have been tuned for speed but their accuracy trails that of two-stage methods. SSD has a 10-20\% lower AP, while YOLO focuses on an even more extreme speed/accuracy trade-off. See Figure~\ref{fig:speed}. Recent work showed that two-stage detectors can be made fast simply by reducing input image resolution and the number of proposals, but one-stage methods trailed in accuracy even with a larger compute budget \cite{Huang2016}. In contrast, the aim of this work is to understand if one-stage detectors can match or surpass the accuracy of two-stage detectors while running at similar or faster speeds.

The design of our RetinaNet detector shares many similarities with previous dense detectors, in particular the concept of `anchors' introduced by RPN \cite{Ren2015a} and use of features pyramids as in SSD \cite{Liu2016} and FPN \cite{Lin2016}. We emphasize that our simple detector achieves top results not based on innovations in network design but due to our novel loss.

\paragraph{Class Imbalance:} Both classic one-stage object detection methods, like boosted detectors \cite{Viola2001, Dollar2009} and DPMs \cite{Felzenszwalb2010a}, and more recent methods, like SSD \cite{Liu2016}, face a large class imbalance during training. These detectors evaluate $10^4$-$10^5$ candidate locations per image but only a few locations contain objects. This imbalance causes two problems: (1) training is inefficient as most locations are easy negatives that contribute no useful learning signal; (2) en masse, the easy negatives can overwhelm training and lead to degenerate models. A common solution is to perform some form of hard negative mining \cite{Sung1994, Viola2001, Felzenszwalb2010a, Shrivastava2016, Liu2016} that samples hard examples during training or more complex sampling/reweighing schemes \cite{Bulo2017}. In contrast, we show that our proposed focal loss naturally handles the class imbalance faced by a one-stage detector and allows us to efficiently train on all examples without sampling and without easy negatives overwhelming the loss and computed gradients.

\paragraph{Robust Estimation:} There has been much interest in designing robust loss functions (\eg, Huber loss \cite{Hastie2008}) that reduce the contribution of \emph{outliers} by down-weighting the loss of examples with large errors (hard examples). In contrast, rather than addressing outliers, our focal loss is designed to address class imbalance by down-weighting \emph{inliers} (easy examples) such that their contribution to the total loss is small even if their number is large. In other words, the focal loss performs the \emph{opposite} role of a robust loss: it focuses training on a sparse set of hard examples.

\section{Focal Loss}\label{sec:loss}

The \emph{Focal Loss} is designed to address the one-stage object detection scenario in which there is an extreme imbalance between foreground and background classes during training (\eg, 1:1000). We introduce the focal loss starting from the cross entropy (CE) loss for binary classification\footnote{Extending the focal loss to the multi-class case is straightforward and works well; for simplicity we focus on the binary loss in this work.}:
 \eqnnm{ce}{\CE(p,y) = \begin{cases} -\log(p) &\text{if $y = 1$}\\
 -\log (1 - p) &\text{otherwise.}\end{cases}}
In the above $y \in \{\pm1\}$ specifies the ground-truth class and $p \in [0,1]$ is the model's estimated probability for the class with label $y=1$. For notational convenience, we define $\pt$:
 \eqnnm{pt}{\pt=\begin{cases} p &\text{if $y = 1$}\\ 1 - p &\text{otherwise,}\end{cases}}
and rewrite $\CE(p,y) = \CE(\pt) = - \log (\pt)$.

The CE loss can be seen as the blue (top) curve in Figure~\ref{fig:loss}. One notable property of this loss, which can be easily seen in its plot, is that even examples that are easily classified ($\pt\gg.5$) incur a loss with non-trivial magnitude. When summed over a large number of easy examples, these small loss values can overwhelm the rare class.

\subsection{Balanced Cross Entropy}\label{sec:loss:alpha}

A common method for addressing class imbalance is to introduce a weighting factor $\alpha \in [0,1]$ for class $1$ and $1 - \alpha$ for class $-1$. In practice $\alpha$ may be set by inverse class frequency or treated as a hyperparameter to set by cross validation. For notational convenience, we define $\at$ analogously to how we defined $\pt$. We write the $\alpha$-balanced CE loss as:
 \eqnnm{cealpha}{\CE(\pt) = - \at \log (\pt).}
This loss is a simple extension to CE that we consider as an experimental baseline for our proposed focal loss.

\subsection{Focal Loss Definition}\label{sec:loss:focal}

As our experiments will show, the large class imbalance encountered during training of dense detectors overwhelms the cross entropy loss. Easily classified negatives comprise the majority of the loss and dominate the gradient. While $\alpha$ balances the importance of positive/negative examples, it does not differentiate between easy/hard examples. Instead, we propose to reshape the loss function to down-weight easy examples and thus focus training on hard negatives.

More formally, we propose to add a modulating factor {\small $(1 - \pt)^\gamma$} to the cross entropy loss, with tunable \emph{focusing} parameter $\gamma \ge 0$. We define the focal loss as:
 \eqnnm{fl}{\FL(\pt) = - (1 - \pt)^\gamma \log (\pt).}

The focal loss is visualized for several values of $\gamma \in [0,5]$ in Figure~\ref{fig:loss}. We note two properties of the focal loss. (1) When an example is misclassified and $\pt$ is small, the modulating factor is near $1$ and the loss is unaffected. As $\pt \rightarrow 1$, the factor goes to 0 and the loss for well-classified examples is down-weighted. (2) The focusing parameter $\gamma$ smoothly adjusts the rate at which easy examples are down-weighted. When $\gamma = 0$, FL is equivalent to CE, and as $\gamma$ is increased the effect of the modulating factor is likewise increased (we found $\gamma = 2$ to work best in our experiments).

Intuitively, the modulating factor reduces the loss contribution from easy examples and extends the range in which an example receives low loss. For instance, with $\gamma = 2$, an example classified with $\pt = 0.9$ would have $100\x$ lower loss compared with CE and with $\pt \approx 0.968$ it would have $1000\x$ lower loss. This in turn increases the importance of correcting misclassified examples (whose loss is scaled down by at most $4\x$ for $\pt\le.5$ and $\gamma=2$).

In practice we use an $\alpha$-balanced variant of the focal loss:
 \eqnnm{flalpha}{\FL(\pt) = - \at (1 - \pt)^\gamma \log (\pt).}
We adopt this form in our experiments as it yields slightly improved accuracy over the non-$\alpha$-balanced form. Finally, we note that the implementation of the loss layer combines the sigmoid operation for computing $p$ with the loss computation, resulting in greater numerical stability.

While in our main experimental results we use the focal loss definition above, its precise form is not crucial. In the appendix we consider other instantiations of the focal loss and demonstrate that these can be equally effective.

\subsection{Class Imbalance and Model Initialization}\label{sec:loss:init}

Binary classification models are by default initialized to have equal probability of outputting either $y=-1$ or $1$. Under such an initialization, in the presence of class imbalance, the loss due to the frequent class can dominate total loss and cause instability in early training. To counter this, we introduce the concept of a `prior' for the value of $p$ estimated by the model for the rare class (foreground) \emph{at the start of training}. We denote the prior by $\pi$ and set it so that the model's estimated $p$ for examples of the rare class is low, \eg $0.01$. We note that this is a change in model initialization (see \S\ref{sec:model:details}) and \emph{not} of the loss function. We found this to improve training stability for both the cross entropy and focal loss in the case of heavy class imbalance.

\subsection{Class Imbalance and Two-stage Detectors}\label{sec:loss:twostage}

Two-stage detectors are often trained with the cross entropy loss without use of $\alpha$-balancing or our proposed loss. Instead, they address class imbalance through two mechanisms: (1) a two-stage cascade and (2) biased minibatch sampling. The first cascade stage is an object proposal mechanism \cite{Uijlings2013, Pinheiro2015, Ren2015a} that reduces the nearly infinite set of possible object locations down to one or two thousand. Importantly, the selected proposals are not random, but are likely to correspond to true object locations, which removes the vast majority of easy negatives. When training the second stage, biased sampling is typically used to construct minibatches that contain, for instance, a 1:3 ratio of positive to negative examples. This ratio is like an implicit $\alpha$-balancing factor that is implemented via sampling. Our proposed focal loss is designed to address these mechanisms in a one-stage detection system directly via the loss function.

\begin{figure*}[t]
\centering
\includegraphics[width=1\linewidth]{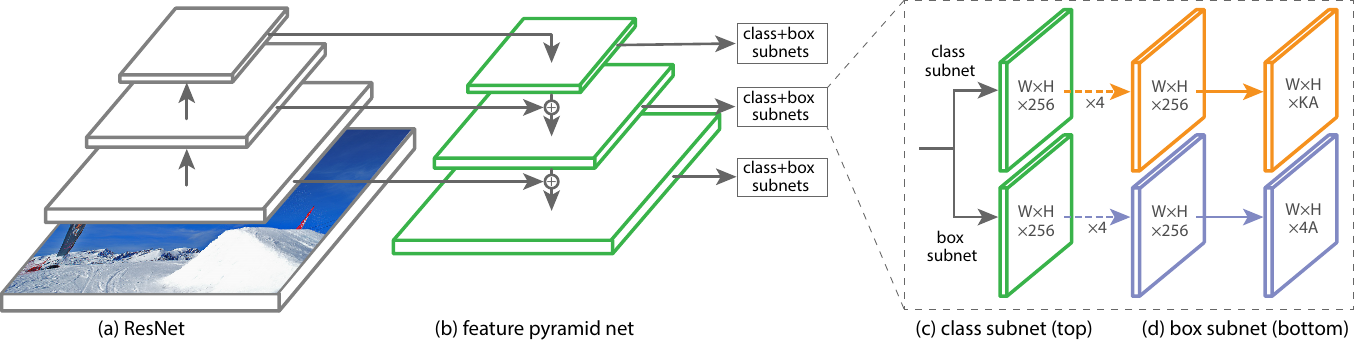}\vspace{1mm}
\caption{The one-stage \bd{RetinaNet} network architecture uses a Feature Pyramid Network (FPN) \cite{Lin2016} backbone on top of a feedforward ResNet architecture \cite{He2016} (a) to generate a rich, multi-scale convolutional feature pyramid (b). To this backbone RetinaNet attaches two subnetworks, one for classifying anchor boxes (c) and one for regressing from anchor boxes to ground-truth object boxes (d). The network design is intentionally simple, which enables this work to focus on a novel focal loss function that eliminates the accuracy gap between our one-stage detector and state-of-the-art two-stage detectors like Faster R-CNN with FPN \cite{Lin2016} while running at faster speeds.}
\label{fig:arch}\vspace{-2mm}
\end{figure*}

\section{RetinaNet Detector}\label{sec:model}

RetinaNet is a single, unified network composed of a \emph{backbone} network and two task-specific \emph{subnetworks}. The backbone is responsible for computing a convolutional feature map over an entire input image and is an off-the-self convolutional network. The first subnet performs convolutional object classification on the backbone's output; the second subnet performs convolutional bounding box regression. The two subnetworks feature a simple design that we propose specifically for one-stage, dense detection, see Figure~\ref{fig:arch}. While there are many possible choices for the details of these components, most design parameters are not particularly sensitive to exact values as shown in the experiments. We describe each component of RetinaNet next.

\paragraph{Feature Pyramid Network Backbone:} We adopt the Feature Pyramid Network (FPN) from \cite{Lin2016} as the backbone network for RetinaNet. In brief, FPN augments a standard convolutional network with a top-down pathway and lateral connections so the network efficiently constructs a rich, multi-scale feature pyramid from a single resolution input image, see Figure~\ref{fig:arch}(a)-(b). Each level of the pyramid can be used for detecting objects at a different scale. FPN improves multi-scale predictions from fully convolutional networks (FCN) \cite{Long2015}, as shown by its gains for RPN \cite{Ren2015a} and DeepMask-style proposals \cite{Pinheiro2015}, as well at two-stage detectors such as Fast R-CNN \cite{Girshick2015a} or Mask R-CNN \cite{He2017}.

Following \cite{Lin2016}, we build FPN on top of the ResNet architecture \cite{He2016}. We construct a pyramid with levels $P_3$ through $P_7$, where $l$ indicates pyramid level ($P_l$ has resolution $2^l$ lower than the input). As in \cite{Lin2016} all pyramid levels have $C=256$ channels. Details of the pyramid generally follow \cite{Lin2016} with a few modest differences.\footnote{RetinaNet uses feature pyramid levels $P_3$ to $P_7$, where $P_3$ to $P_5$ are computed from the output of the corresponding ResNet residual stage ($C_3$ through $C_5$) using top-down and lateral connections just as in \cite{Lin2016}, $P_6$ is obtained via a 3$\x$3 stride-2 conv on $C_5$, and $P_7$ is computed by applying ReLU followed by a 3$\x$3 stride-2 conv on $P_6$. This differs slightly from \cite{Lin2016}: (1) we don't use the high-resolution pyramid level $P_2$ for computational reasons, (2) $P_6$ is computed by strided convolution instead of downsampling, and (3) we include $P_7$ to improve large object detection. These minor modifications improve speed while maintaining accuracy.} While many design choices are not crucial, we emphasize the use of the FPN backbone is; preliminary experiments using features from only the final ResNet layer yielded low AP.

\paragraph{Anchors:} We use translation-invariant anchor boxes similar to those in the RPN variant in \cite{Lin2016}. The anchors have areas of $32^2$ to $512^2$ on pyramid levels $P_3$ to $P_7$, respectively. As in \cite{Lin2016}, at each pyramid level we use anchors at three aspect ratios $\{1$:$2,$ $1$:$1$, $2$:$1\}$. For denser scale coverage than in \cite{Lin2016}, at each level we add anchors of sizes {\small \{$2^{0}$, $2^{1/3}$, $2^{2/3}$\}} of the original set of 3 aspect ratio anchors. This improve AP in our setting. In total there are $A=9$ anchors per level and across levels they cover the scale range 32 - 813 pixels with respect to the network's input image.

Each anchor is assigned a length $K$ one-hot vector of classification targets, where $K$ is the number of object classes, and a 4-vector of box regression targets. We use the assignment rule from RPN \cite{Ren2015a} but modified for multi-class detection and with adjusted thresholds. Specifically, anchors are assigned to ground-truth object boxes using an intersection-over-union (IoU) threshold of 0.5; and to background if their IoU is in [0, 0.4). As each anchor is assigned to at most one object box, we set the corresponding entry in its length $K$ label vector to $1$ and all other entries to $0$. If an anchor is unassigned, which may happen with overlap in [0.4, 0.5), it is ignored during training. Box regression targets are computed as the offset between each anchor and its assigned object box, or omitted if there is no assignment.

\paragraph{Classification Subnet:} The classification subnet predicts the probability of object presence at each spatial position for each of the $A$ anchors and $K$ object classes. This subnet is a small FCN attached to each FPN level; parameters of this subnet are shared across all pyramid levels. Its design is simple. Taking an input feature map with $C$ channels from a given pyramid level, the subnet applies four 3$\x$3 conv layers, each with $C$ filters and each followed by ReLU activations, followed by a 3$\x$3 conv layer with $KA$ filters. Finally sigmoid activations are attached to output the $KA$ binary predictions per spatial location, see Figure~\ref{fig:arch} (c). We use $C = 256$ and $A = 9$ in most experiments.

In contrast to RPN \cite{Ren2015a}, our object classification subnet is deeper, uses only 3$\x$3 convs, and does not share parameters with the box regression subnet (described next). We found these higher-level design decisions to be more important than specific values of hyperparameters.

\paragraph{Box Regression Subnet:} In parallel with the object classification subnet, we attach another small FCN to each pyramid level for the purpose of regressing the offset from each anchor box to a nearby ground-truth object, if one exists. The design of the box regression subnet is identical to the classification subnet except that it terminates in $4A$ linear outputs per spatial location, see Figure~\ref{fig:arch} (d). For each of the $A$ anchors per spatial location, these $4$ outputs predict the relative offset between the anchor and the ground-truth box (we use the standard box parameterization from R-CNN \cite{Girshick2014}). We note that unlike most recent work, we use a class-agnostic bounding box regressor which uses fewer parameters and we found to be equally effective. The object classification subnet and the box regression subnet, though sharing a common structure, use separate parameters.

\subsection{Inference and Training}\label{sec:model:details}

\paragraph{Inference:} RetinaNet forms a single FCN comprised of a ResNet-FPN backbone, a classification subnet, and a box regression subnet, see Figure~\ref{fig:arch}. As such, inference involves simply forwarding an image through the network. To improve speed, we only decode box predictions from at most 1k top-scoring  predictions per FPN level, after thresholding detector confidence at 0.05. The top predictions from all levels are merged and non-maximum suppression with a threshold of 0.5 is applied to yield the final detections.

\paragraph{Focal Loss:} We use the focal loss introduced in this work as the loss on the output of the classification subnet. As we will show in \S\ref{sec:exps}, we find that $\gamma = 2$ works well in practice and the RetinaNet is relatively robust to $\gamma \in [0.5, 5]$. We emphasize that when training RetinaNet, the focal loss is applied to \emph{all} \app100k anchors in each sampled image. This stands in contrast to common practice of using heuristic sampling (RPN) or hard example mining (OHEM, SSD) to select a small set of anchors (\eg, 256) for each minibatch. The total focal loss of an image is computed as the sum of the focal loss over all \app100k anchors, \emph{normalized by the number of anchors assigned to a ground-truth box}. We perform the normalization by the number of assigned anchors, not total anchors, since the vast majority of anchors are easy negatives and receive negligible loss values under the focal loss. Finally we note that $\alpha$, the weight assigned to the rare class, also has a stable range, but it interacts with $\gamma$ making it necessary to select the two together (see Tables~\ref{tab:ablation:alpha} and \ref{tab:ablation:gamma}). In general $\alpha$ should be decreased slightly as $\gamma$ is increased (for $\gamma=2$, $\alpha=0.25$ works best).

\begin{table*}[t]\centering
\subfloat[\bd{Varying $\alpha$ for CE loss} ($\gamma = 0$)\label{tab:ablation:alpha}]{
\tablestyle{4pt}{1.1}\begin{tabular}{c|x{22}x{22}x{22}}
 $\alpha$ & AP & AP$_{50}$ & AP$_{75}$\\[.1em]
 \shline
 .10  &  0.0 &  0.0 &  0.0 \\
 .25  & 10.8 & 16.0 & 11.7 \\
 .50  & 30.2 & 46.7 & 32.8 \\
 .75  & 31.1 & 49.4 & 33.0 \\
 .90  & 30.8 & 49.7 & 32.3 \\
 .99  & 28.7 & 47.4 & 29.9 \\
 .999 & 25.1 & 41.7 & 26.1 \\
\end{tabular}}\hspace{8mm}
\subfloat[\bd{Varying $\gamma$ for FL} (w. optimal $\alpha$)\label{tab:ablation:gamma}]{
\tablestyle{4pt}{1.1}\begin{tabular}{cc|x{22}x{22}x{22}}
 $\gamma$ & $\alpha$ & AP & AP$_{50}$ & AP$_{75}$\\[.1em]
 \shline
 0   & .75 & 31.1 & 49.4 & 33.0\\
 0.1 & .75 & 31.4 & 49.9 & 33.1\\
 0.2 & .75 & 31.9 & 50.7 & 33.4\\
 0.5 & .50 & 32.9 & 51.7 & 35.2\\
 1.0 & .25 & 33.7 & 52.0 & 36.2\\
 2.0 & .25 & \bd{34.0} & \bd{52.5} & \bd{36.5}\\
 5.0 & .25 & 32.2 & 49.6 & 34.8\\
\end{tabular}}\hspace{8mm}
\subfloat[\bd{Varying anchor scales and aspects}\label{tab:ablation:anchors}]{
\tablestyle{4pt}{1.1}\begin{tabular}{cc|x{22}x{22}x{22}}
 \#sc & \#ar &
 AP & AP$_{50}$ & AP$_{75}$\\[.1em]
 \shline
 1 & 1 & 30.3 & 49.0 & 31.8\\
 2 & 1 & 31.9 & 50.0 & 34.0\\
 3 & 1 & 31.8 & 49.4 & 33.7\\
 1 & 3 & 32.4 & 52.3 & 33.9\\
 2 & 3 & \bd{34.2} & \bd{53.1} & \bd{36.5}\\
 3 & 3 & 34.0 & 52.5 & \bd{36.5}\\
 4 & 3 & 33.8 & 52.1 & 36.2\\
\end{tabular}}\\[-4pt]
\subfloat[\bd{FL \vs OHEM} baselines (with ResNet-101-FPN)\label{tab:ablation:ohem}]{
\tablestyle{4pt}{1.1}\begin{tabular}{c|cc|x{22}x{22}x{22}}
 \mrtwo{method} & batch & nms & \mrtwo{AP} & \mrtwo{AP$_{50}$} & \mrtwo{AP$_{75}$}\\[-3pt]
 & size  & thr  & \\
 \shline
 OHEM     & 128 & .7 & 31.1 & 47.2 & 33.2 \\
 OHEM     & 256 & .7 & 31.8 & 48.8 & 33.9 \\
 OHEM     & 512 & .7 & 30.6 & 47.0 & 32.6 \\
 \hline
 OHEM     & 128 & .5 & 32.8 & 50.3 & 35.1 \\
 OHEM     & 256 & .5 & 31.0 & 47.4 & 33.0 \\
 OHEM     & 512 & .5 & 27.6 & 42.0 & 29.2 \\
 \hline
 OHEM 1:3 & 128 & .5 & 31.1 & 47.2 & 33.2 \\
 OHEM 1:3 & 256 & .5 & 28.3 & 42.4 & 30.3 \\
 OHEM 1:3 & 512 & .5 & 24.0 & 35.5 & 25.8 \\
 \hline
 \bd{FL} & n/a & n/a & \bd{36.0} & \bd{54.9} & \bd{38.7} \\
\end{tabular}}\hspace{3mm}
\subfloat[\bd{Accuracy/speed trade-off} RetinaNet (on \symb{test-dev})\label{tab:ablation:models}]{
\tablestyle{3pt}{1.1}\begin{tabular}{cc|x{22}x{22}x{22}|x{22}x{22}x{22}|c}\\[-.8em]
 depth & scale & AP & AP$_{50}$ & AP$_{75}$ & AP$_S$ & AP$_M$ & AP$_L$ & time \\[.4em]
 \shline
 50 & 400  & 30.5 & 47.8 & 32.7 & 11.2 & 33.8 & 46.1 &  64 \\
 50 & 500  & 32.5 & 50.9 & 34.8 & 13.9 & 35.8 & 46.7 &  72 \\
 50 & 600  & 34.3 & 53.2 & 36.9 & 16.2 & 37.4 & 47.4 &  98 \\
 50 & 700  & 35.1 & 54.2 & 37.7 & 18.0 & 39.3 & 46.4 & 121 \\
 50 & 800  & 35.7 & 55.0 & 38.5 & 18.9 & 38.9 & 46.3 & 153 \\
 \hline
 101 & 400 & 31.9 & 49.5 & 34.1 & 11.6 & 35.8 & 48.5 &  81 \\
 101 & 500 & 34.4 & 53.1 & 36.8 & 14.7 & 38.5 & 49.1 &  90 \\
 101 & 600 & 36.0 & 55.2 & 38.7 & 17.4 & 39.6 & 49.7 & 122 \\
 101 & 700 & 37.1 & 56.6 & 39.8 & 19.1 & 40.6 & 49.4 & 154 \\
 101 & 800 & 37.8 & 57.5 & 40.8 & 20.2 & 41.1 & 49.2 & 198 \\
\end{tabular}}\vspace{2mm}
\caption{\textbf{Ablation experiments for RetinaNet and Focal Loss (FL).} All models are trained on \symb{trainval35k} and tested on \symb{minival} unless noted. If not specified, default values are: $\gamma = 2$; anchors for 3 scales and 3 aspect ratios; ResNet-50-FPN backbone; and a 600 pixel train and test image scale. (a) RetinaNet with $\alpha$-balanced CE achieves at most 31.1 AP. (b) In contrast, using FL with the same exact network gives a 2.9 AP gain and is fairly robust to exact $\gamma$/$\alpha$ settings. (c) Using 2-3 scale and 3 aspect ratio anchors yields good results after which point performance saturates. (d) FL outperforms the best variants of online hard example mining (OHEM) \cite{Shrivastava2016,Liu2016} by over 3 points AP. (e) Accuracy/Speed trade-off of RetinaNet on \symb{test-dev} for various network depths and image scales (see also Figure \ref{fig:speed}).}
\label{tab:ablations}
\end{table*}

\paragraph{Initialization:} We experiment with ResNet-50-FPN and ResNet-101-FPN backbones \cite{Lin2016}. The base ResNet-50 and ResNet-101 models are pre-trained on ImageNet1k; we use the models released by \cite{He2016}. New layers added for FPN are initialized as in \cite{Lin2016}. All new conv layers except the final one in the RetinaNet subnets are initialized with bias $b = 0$ and a Gaussian weight fill with $\sigma = 0.01$. For the final conv layer of the classification subnet, we set the bias initialization to $b = -\log((1-\pi)/\pi)$, where $\pi$ specifies that at the start of training every anchor should be labeled as foreground with confidence of \app$\pi$. We use $\pi=.01$ in all experiments, although results are robust to the exact value. As explained in \S\ref{sec:loss:init}, this initialization prevents the large number of background anchors from generating a large, destabilizing loss value in the first iteration of training.

\paragraph{Optimization:} RetinaNet is trained with stochastic gradient descent (SGD). We use synchronized SGD over 8 GPUs with a total of 16 images per minibatch (2 images per GPU). Unless otherwise specified, all models are trained for 90k iterations with an initial learning rate of 0.01, which is then divided by 10 at 60k and again at 80k iterations.  We use horizontal image flipping as the only form of data augmentation unless otherwise noted. Weight decay of 0.0001 and momentum of 0.9 are used. The training loss is the sum the focal loss and the standard smooth $L_1$ loss used for box regression \cite{Girshick2015a}. Training time ranges between 10 and 35 hours for the models in Table~\ref{tab:ablation:models}.

\section{Experiments}\label{sec:exps}

We present experimental results on the bounding box detection track of the challenging COCO benchmark \cite{Lin2014}. For training, we follow common practice \cite{Bell2016,Lin2016} and use the COCO \symb{trainval35k} split (union of 80k images from \symb{train} and a random 35k subset of images from the 40k image \symb{val} split). We report lesion and sensitivity studies by evaluating on the \symb{minival} split (the remaining 5k images from \symb{val}). For our main results, we report COCO AP on the \symb{test-dev} split, which has no public labels and requires use of the evaluation server.

\begin{figure*}[t]\centering
\includegraphics[width=.43\linewidth]{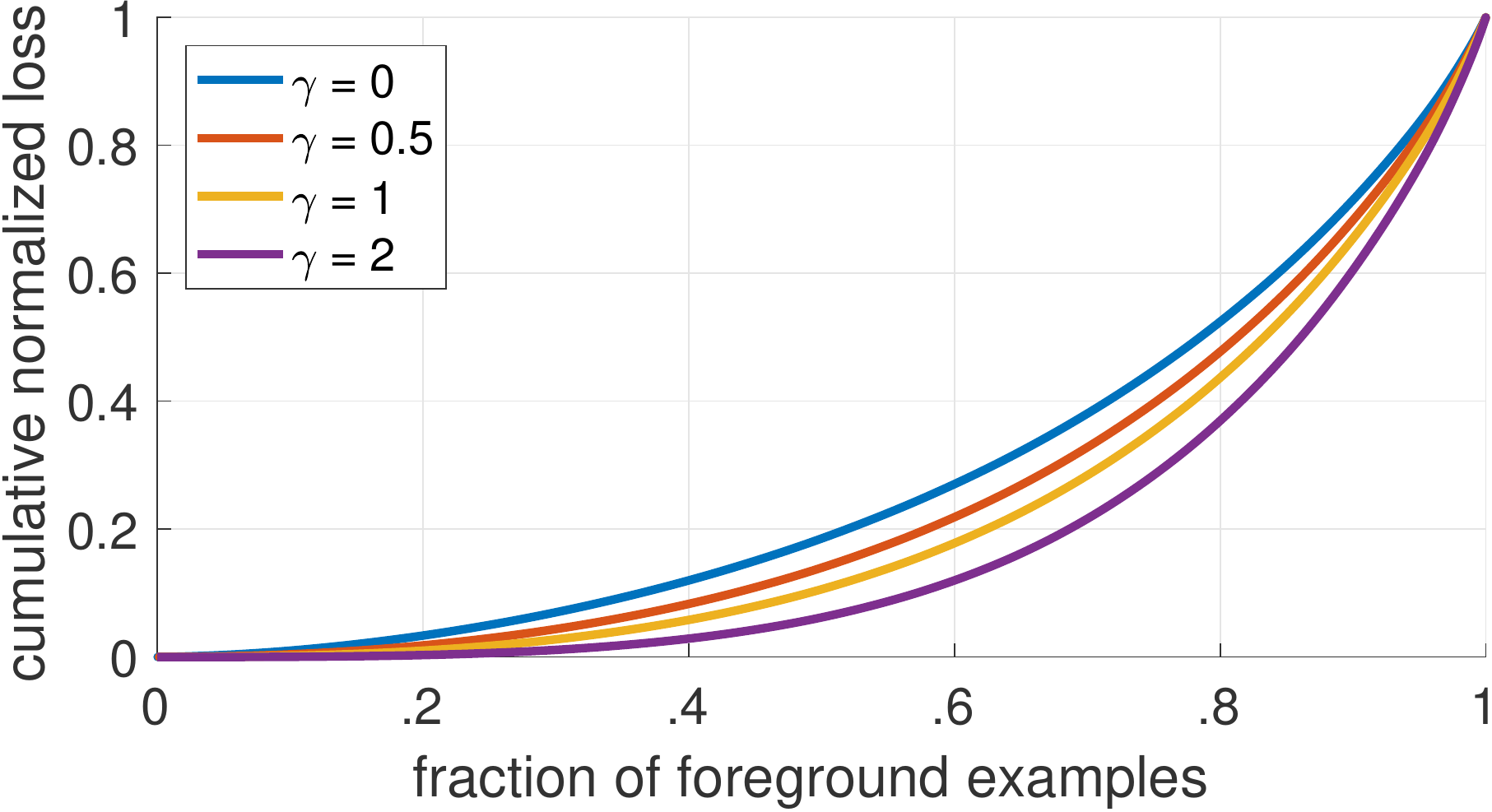}
\hspace{10mm}
\includegraphics[width=.43\linewidth]{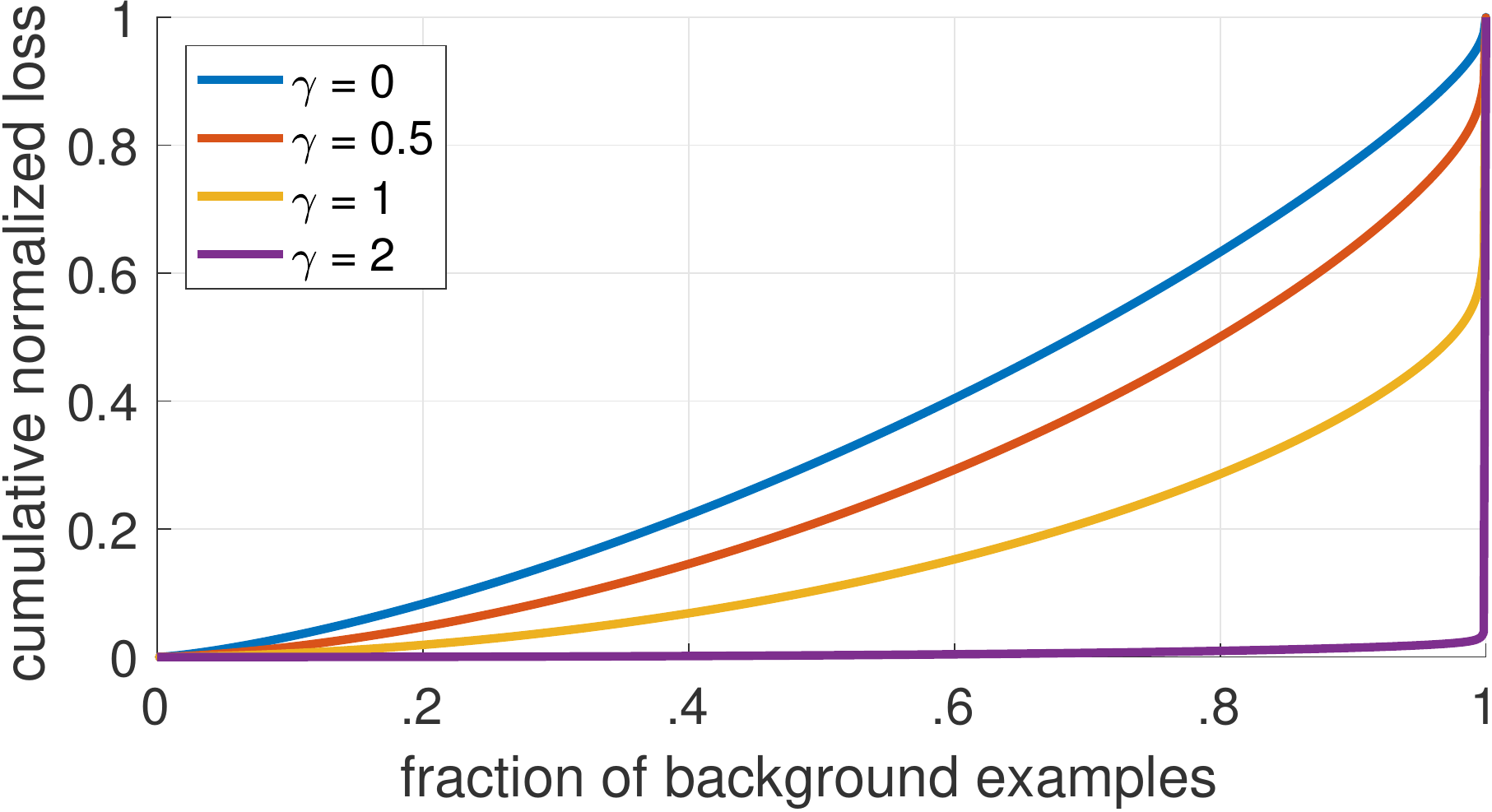}
\caption{Cumulative distribution functions of the normalized loss for positive and negative samples for different values of $\gamma$ for a \emph{converged} model. The effect of changing $\gamma$ on the distribution of the loss for positive examples is minor. For negatives, however, increasing $\gamma$ heavily concentrates the loss on hard examples, focusing nearly all attention away from easy negatives. }
\label{fig:cdfs}\vspace{-2mm}
\end{figure*}

\subsection{Training Dense Detection}\label{sec:exps:training}

We run numerous experiments to analyze the behavior of the loss function for dense detection along with various optimization strategies. For all experiments we use depth 50 or 101 ResNets \cite{He2016} with a Feature Pyramid Network (FPN)~\cite{Lin2016} constructed on top. For all ablation studies we use an image scale of 600 pixels for training and testing.

\paragraph{Network Initialization:} Our first attempt to train RetinaNet uses standard cross entropy (CE) loss without any modifications to the initialization or learning strategy. This fails quickly, with the network diverging during training. However, simply initializing the last layer of our model such that the prior probability of detecting an object is $\pi=.01$ (see \S\ref{sec:model:details}) enables effective learning. Training RetinaNet with ResNet-50 and this initialization already yields a respectable AP of 30.2 on COCO. Results are insensitive to the exact value of $\pi$ so we use $\pi=.01$ for all experiments.

\paragraph{Balanced Cross Entropy:} Our next attempt to improve learning involved using the $\alpha$-balanced CE loss described in \S\ref{sec:loss:alpha}. Results for various $\alpha$ are shown in Table~\ref{tab:ablation:alpha}. Setting $\alpha=.75$ gives a gain of 0.9 points AP.

\paragraph{Focal Loss:} Results using our proposed focal loss are shown in Table~\ref{tab:ablation:gamma}. The focal loss introduces one new hyperparameter, the focusing parameter $\gamma$, that controls the strength of the modulating term. When $\gamma = 0$, our loss is equivalent to the CE loss. As $\gamma$ increases, the shape of the loss changes so that ``easy'' examples with low loss get further discounted, see Figure~\ref{fig:loss}. FL shows large gains over CE as $\gamma$ is increased. With $\gamma=2$, FL \emph{yields a 2.9 AP improvement over the $\alpha$-balanced CE loss}.

For the experiments in Table~\ref{tab:ablation:gamma}, for a fair comparison we find the best $\alpha$ for each $\gamma$. We observe that lower $\alpha$'s are selected for higher $\gamma$'s (as easy negatives are down-weighted, less emphasis needs to be placed on the positives). Overall, however, the benefit of changing $\gamma$ is much larger, and indeed the best $\alpha$'s ranged in just [.25,.75] (we tested $\alpha\in[.01,.999]$). We use $\gamma=2.0$ with $\alpha=.25$ for all experiments but $\alpha=.5$ works nearly as well (.4 AP lower).

\paragraph{Analysis of the Focal Loss:} To understand the focal loss better, we analyze the empirical distribution of the loss of a \emph{converged} model. For this, we take take our default ResNet-101 600-pixel model trained with $\gamma=2$ (which has 36.0 AP). We apply this model to a large number of random images and sample the predicted probability for \app$10^7$ negative windows and \app$10^5$ positive windows. Next, separately for positives and negatives, we compute FL for these samples, and normalize the loss such that it sums to one. Given the normalized loss, we can sort the loss from lowest to highest and plot its cumulative distribution function (CDF) for both positive and negative samples and for different settings for $\gamma$ (even though model was trained with $\gamma=2$).

Cumulative distribution functions for positive and negative samples are shown in Figure~\ref{fig:cdfs}. If we observe the positive samples, we see that the CDF looks fairly similar for different values of $\gamma$. For example, approximately 20\% of the hardest positive samples account for roughly half of the positive loss, as $\gamma$ increases more of the loss gets concentrated in the top 20\% of examples, but the effect is minor.

The effect of $\gamma$ on negative samples is dramatically different. For $\gamma=0$, the positive and negative CDFs are quite similar. However, as $\gamma$ increases, substantially more weight becomes concentrated on the hard negative examples. In fact, with $\gamma=2$ (our default setting), the vast majority of the loss comes from a small fraction of samples. As can be seen, FL can effectively discount the effect of easy negatives, focusing all attention on the hard negative examples.

\begin{table*}[t]
\tablestyle{4pt}{1.05}
\begin{tabular}{l|c|x{22}x{22}x{22}|x{22}x{22}x{22}}
 & backbone
 & AP & AP$_{50}$ & AP$_{75}$
 & AP$_S$ & AP$_M$ &  AP$_L$\\ [.1em]
\shline
\emph{Two-stage methods} & & & & & & & \\
 ~Faster R-CNN+++ \cite{He2016} & ResNet-101-C4
  & 34.9 & 55.7 & 37.4 & 15.6 & 38.7 & 50.9\\
 ~Faster R-CNN w FPN \cite{Lin2016} & ResNet-101-FPN
  & 36.2 & 59.1 & 39.0 & 18.2 & 39.0 & 48.2\\
 ~Faster R-CNN by G-RMI \cite{Huang2016} & Inception-ResNet-v2 \cite{Szegedy2016a}
  & 34.7 & 55.5 & 36.7 & 13.5 & 38.1 & 52.0\\
 ~Faster R-CNN w TDM \cite{Shrivastava2016a} & Inception-ResNet-v2-TDM
  & 36.8 & 57.7 & 39.2 & 16.2 & 39.8 & \bd{52.1}\\
\hline
\emph{One-stage methods} & & & & & & & \\
 ~YOLOv2 \cite{Redmon2017} & DarkNet-19 \cite{Redmon2017}
  & 21.6 & 44.0 & 19.2 & 5.0 & 22.4 & 35.5 \\
 ~SSD513 \cite{Liu2016,Fu2016} & ResNet-101-SSD
  & 31.2 & 50.4 & 33.3 & 10.2 & 34.5 & 49.8 \\
 ~DSSD513 \cite{Fu2016} & ResNet-101-DSSD
  & 33.2 & 53.3 & 35.2 & 13.0 & 35.4 & 51.1 \\
 ~\bd{RetinaNet} (ours) & ResNet-101-FPN
  & 39.1 & 59.1 & 42.3 & 21.8 & 42.7 & 50.2 \\
 ~\bd{RetinaNet} (ours) & ResNeXt-101-FPN
  & \bd{40.8} & \bd{61.1} & \bd{44.1} & \bd{24.1} & \bd{44.2} & 51.2 \\
\end{tabular}\vspace{2mm}
\caption{\textbf{Object detection} \emph{single-model} results (bounding box AP), \vs state-of-the-art on COCO \symb{test-dev}. We show results for our RetinaNet-101-800 model, trained  with scale jitter and for 1.5$\x$ longer than the same model from Table~\ref{tab:ablation:models}. Our model achieves top results, outperforming both one-stage and two-stage models. For a detailed breakdown of speed versus accuracy see Table~\ref{tab:ablation:models} and Figure~\ref{fig:speed}.}
\label{tab:final_bbox}\vspace{-3mm}
\end{table*}

\paragraph{Online Hard Example Mining (OHEM):} \cite{Shrivastava2016} proposed to improve training of two-stage detectors by constructing minibatches using high-loss examples. Specifically, in OHEM each example is scored by its loss, non-maximum suppression (nms) is then applied, and a minibatch is constructed with the highest-loss examples. The nms threshold and batch size are tunable parameters. Like the focal loss, OHEM puts more emphasis on misclassified examples, but unlike FL, OHEM completely discards easy examples. We also implement a variant of OHEM used in SSD \cite{Liu2016}: after applying nms to all examples, the minibatch is constructed to enforce a 1:3 ratio between positives and negatives to help ensure each minibatch has enough positives.

We test both OHEM variants in our setting of one-stage detection which has large class imbalance. Results for the original OHEM strategy and the `OHEM 1:3' strategy for selected batch sizes and nms thresholds are shown in Table~\ref{tab:ablation:ohem}. These results use ResNet-101, our baseline trained with FL achieves 36.0 AP for this setting. In contrast, the best setting for OHEM (no 1:3 ratio, batch size 128, nms of .5) achieves 32.8 AP. This is a gap of 3.2 AP, showing FL is more effective than OHEM for training dense detectors. We note that we tried other parameter setting and variants for OHEM but did not achieve better results.

\paragraph{Hinge Loss:} Finally, in early experiments, we attempted to train with the hinge loss \cite{Hastie2008} on $\pt$, which sets loss to 0 above a certain value of $\pt$. However, this was unstable and we did not manage to obtain meaningful results. Results exploring alternate loss functions are in the appendix.

\subsection{Model Architecture Design}\label{sec:exps:model}

\paragraph{Anchor Density:} One of the most important design factors in a one-stage detection system is how densely it covers the space of possible image boxes. Two-stage detectors can classify boxes at any position, scale, and aspect ratio using a region pooling operation \cite{Girshick2015a}. In contrast, as one-stage detectors use a fixed sampling grid, a popular approach for achieving high coverage of boxes in these approaches is to use multiple `anchors' \cite{Ren2015a} at each spatial position to cover boxes of various scales and aspect ratios.

We sweep over the number of scale and aspect ratio anchors used at each spatial position and each pyramid level in FPN. We consider cases from a single square anchor at each location to 12 anchors per location spanning 4 sub-octave scales ({\small $2^{k/4}$}, for $k \le 3$) and 3 aspect ratios [0.5, 1, 2]. Results using ResNet-50 are shown in Table~\ref{tab:ablation:anchors}. A surprisingly good AP (30.3) is achieved using just one square anchor. However, the AP can be improved by nearly 4 points (to 34.0) when using 3 scales and 3 aspect ratios per location. We used this setting for all other experiments in this work.

Finally, we note that increasing beyond 6-9 anchors did not shown further gains. Thus while two-stage systems can classify arbitrary boxes in an image, the saturation of performance \wrt density implies the higher potential density of two-stage systems may not offer an advantage.

\paragraph{Speed versus Accuracy:} Larger backbone networks yield higher accuracy, but also slower inference speeds. Likewise for input image scale (defined by the shorter image side). We show the impact of these two factors in Table~\ref{tab:ablation:models}. In Figure~\ref{fig:speed} we plot the speed/accuracy trade-off curve for RetinaNet and compare it to recent methods using public numbers on COCO \symb{test-dev}. The plot reveals that RetinaNet, enabled by our focal loss, forms an upper envelope over all existing methods, discounting the low-accuracy regime. RetinaNet with ResNet-101-FPN and a 600 pixel image scale (which we denote by RetinaNet-101-600 for simplicity) matches the accuracy of the recently published ResNet-101-FPN Faster R-CNN \cite{Lin2016}, while running in 122 ms per image compared to 172 ms (both measured on an Nvidia M40 GPU). Using larger scales allows RetinaNet to surpass the accuracy of all two-stage approaches, while still being faster. For faster runtimes, there is only one operating point (500 pixel input) at which using ResNet-50-FPN improves over ResNet-101-FPN. Addressing the high frame rate regime will likely require special network design, as in \cite{Redmon2017}, and is beyond the scope of this work.  We note that after publication, faster and more accurate results can now be obtained by a variant of Faster R-CNN from \cite{Detectron2018}.

\subsection{Comparison to State of the Art}\label{sec:exps:results}

We evaluate RetinaNet on the challenging COCO dataset and compare \symb{test-dev} results to recent state-of-the-art methods including both one-stage and two-stage models. Results are presented in Table~\ref{tab:final_bbox} for our RetinaNet-101-800 model trained using scale jitter and for 1.5$\x$ longer than the models in Table~\ref{tab:ablation:models} (giving a 1.3 AP gain). Compared to existing one-stage methods, our approach achieves a healthy 5.9 point AP gap (39.1 \vs 33.2) with the closest competitor, DSSD \cite{Fu2016}, while also being faster, see Figure~\ref{fig:speed}. Compared to recent two-stage methods, RetinaNet achieves a 2.3 point gap above the top-performing Faster R-CNN model based on Inception-ResNet-v2-TDM \cite{Shrivastava2016a}. Plugging in ResNeXt-32x8d-101-FPN \cite{Xie2016} as the RetinaNet backbone further improves results another 1.7 AP, surpassing 40 AP on COCO.

\section{Conclusion}

In this work, we identify class imbalance as the primary obstacle preventing one-stage object detectors from surpassing top-performing, two-stage methods. To address this, we propose the \emph{focal loss} which applies a modulating term to the cross entropy loss in order to focus learning on hard negative examples. Our approach is simple and highly effective. We demonstrate its efficacy by designing a fully convolutional one-stage detector and report extensive experimental analysis showing that it achieves state-of-the-art accuracy and speed. Source code is available at {\footnotesize\url{https://github.com/facebookresearch/Detectron}} \cite{Detectron2018}.

\section*{Appendix A: Focal Loss*}

\begin{figure}[t]\centering
\includegraphics[width=.99\linewidth]{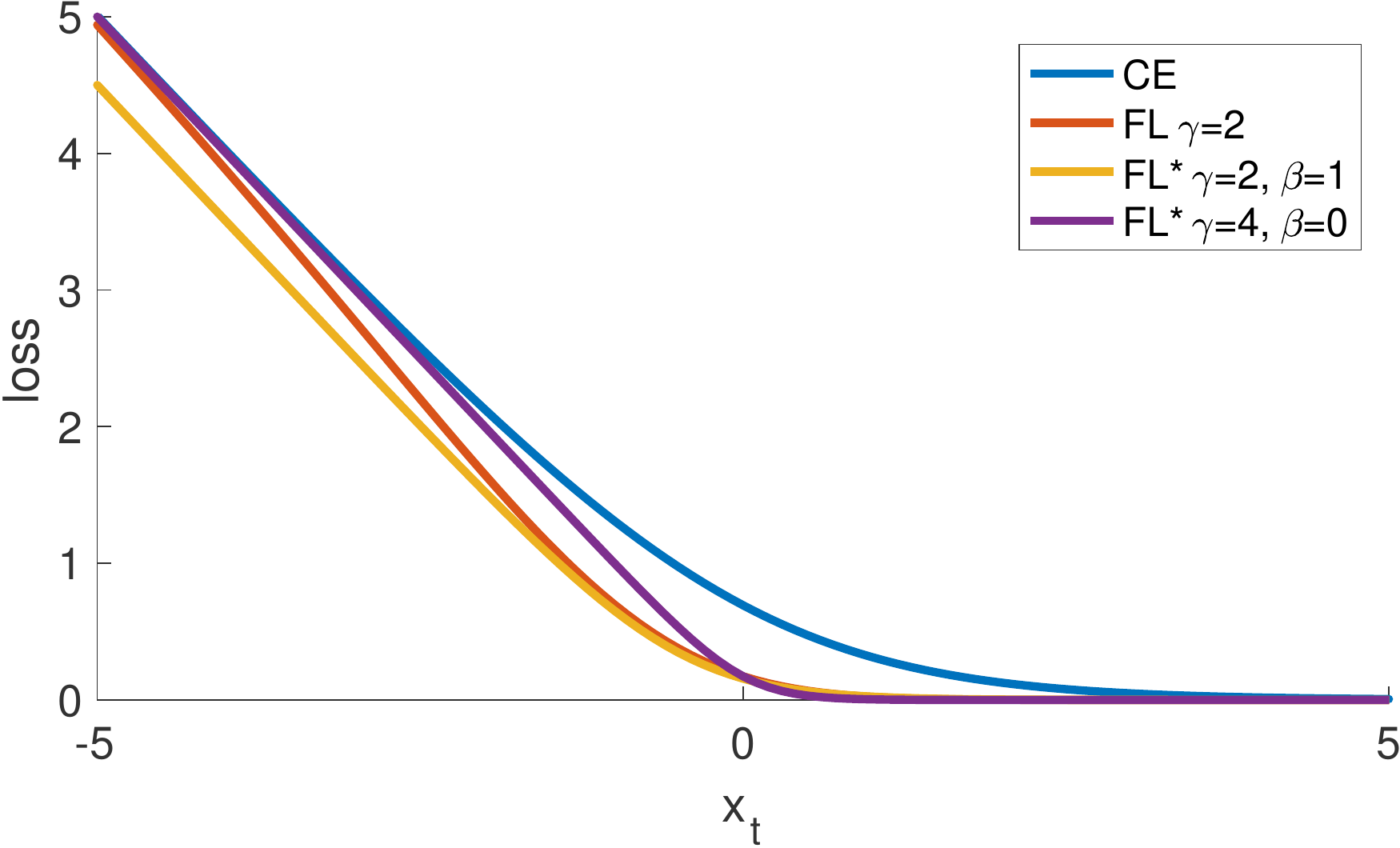}
\caption{Focal loss variants compared to the cross entropy as a function of $\xt=yx$. Both the original FL and alternate variant $\FQ$ reduce the relative loss for well-classified examples ($\xt>0$).}
\label{fig:loss-variants}
\end{figure}

\begin{table}[t]
\tablestyle{4pt}{1.1}\begin{tabular}{c|cc|x{22}x{22}x{22}}
 loss & $\gamma$ & $\beta$ & AP & AP$_{50}$ & AP$_{75}$\\[.1em]
 \shline
 $\CE$ & --  & --  & 31.1 & 49.4 & 33.0\\
 $\FL$ & 2.0 & --  & 34.0 & 52.5 & 36.5\\
 $\FQ$ & 2.0 & 1.0 & 33.8 & 52.7 & 36.3\\
 $\FQ$ & 4.0 & 0.0 & 33.9 & 51.8 & 36.4\\
\end{tabular}
\vspace{2mm}
\caption{Results of FL and $\FQ$ versus CE for select settings.}
\label{tab:FQ}
\end{table}

The exact form of the focal loss is not crucial. We now show an alternate instantiation of the focal loss that has similar properties and yields comparable results. The following also gives more insights into properties of the focal loss.

We begin by considering both cross entropy (CE) and the focal loss (FL) in a slightly different form than in the main text. Specifically, we define a quantity $\xt$ as follows:
 \eqnnm{xt}{\xt = yx,}
where $y \in \{\pm1\}$ specifies the ground-truth class as before. We can then write $\pt=\sigma(\xt)$ (this is compatible with the definition of $\pt$ in Equation \ref{eq:pt}). An example is correctly classified when $\xt>0$, in which case $\pt>.5$.

We can now define an alternate form of the focal loss in terms of $\xt$. We define $\pt^*$ and $\FQ$ as follows:
\begin{gather}
 \pt^* = \sigma(\gamma \xt+\beta),\\
 \FQ = -\log(\pt^*)/\gamma.
\end{gather}
$\FQ$ has two parameters, $\gamma$ and $\beta$, that control the steepness and shift of the loss curve. We plot $\FQ$ for two selected settings of $\gamma$ and $\beta$ in Figure~\ref{fig:loss-variants} alongside CE and FL. As can be seen, like FL, $\FQ$ with the selected parameters diminishes the loss assigned to well-classified examples.

We trained RetinaNet-50-600 using identical settings as before but we swap out FL for $\FQ$ with the selected parameters. These models achieve nearly the same AP as those trained with FL, see Table~\ref{tab:FQ}. In other words, $\FQ$ is a reasonable alternative for the FL that works well in practice.

\begin{figure}[t]\centering
\includegraphics[width=.99\linewidth]{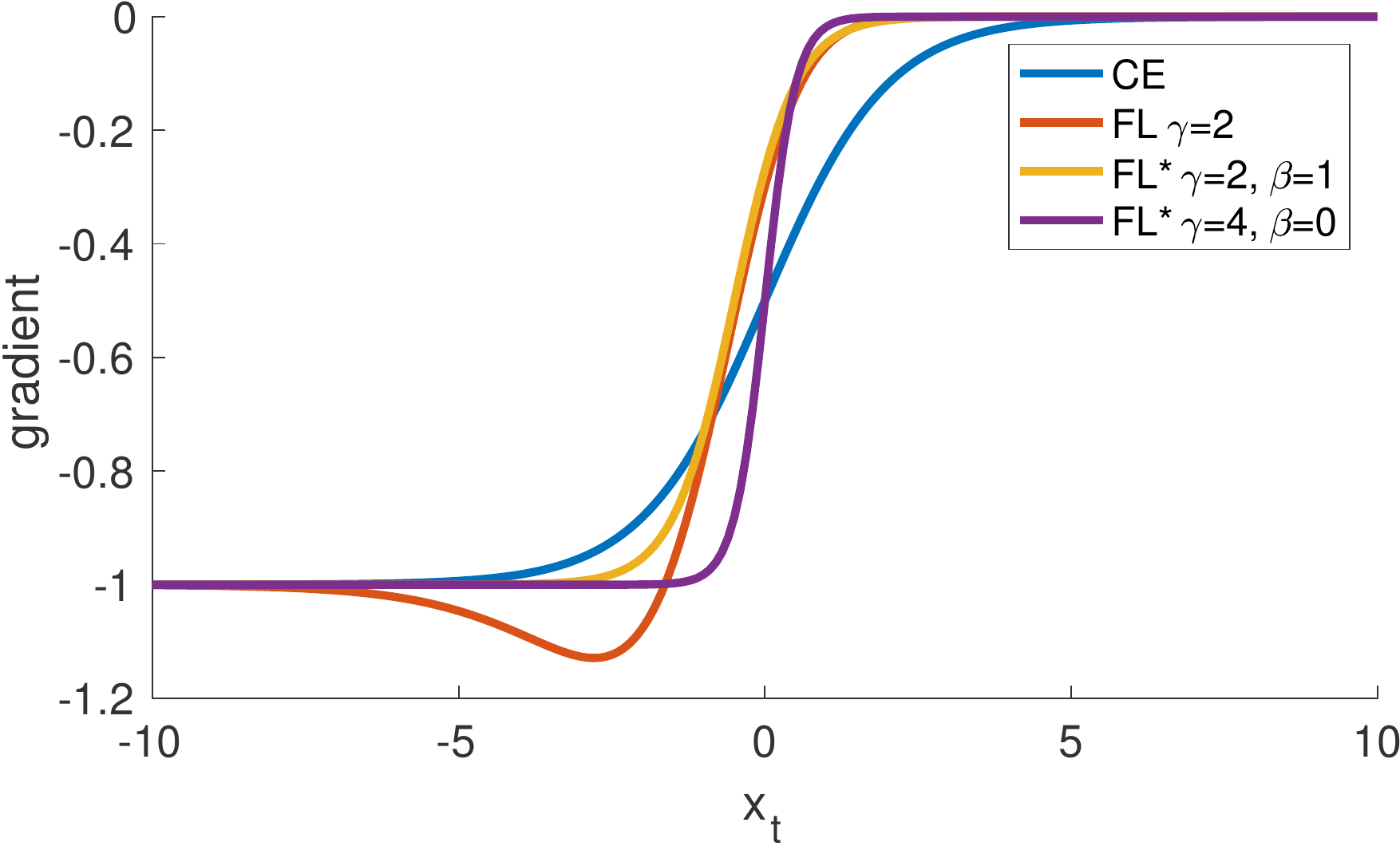}
\caption{Derivates of the loss functions from Figure~\ref{fig:loss-variants} \wrt $x$.}
\label{fig:loss-derivatives}
\end{figure}

\begin{figure}[t]\centering
\includegraphics[width=.99\linewidth]{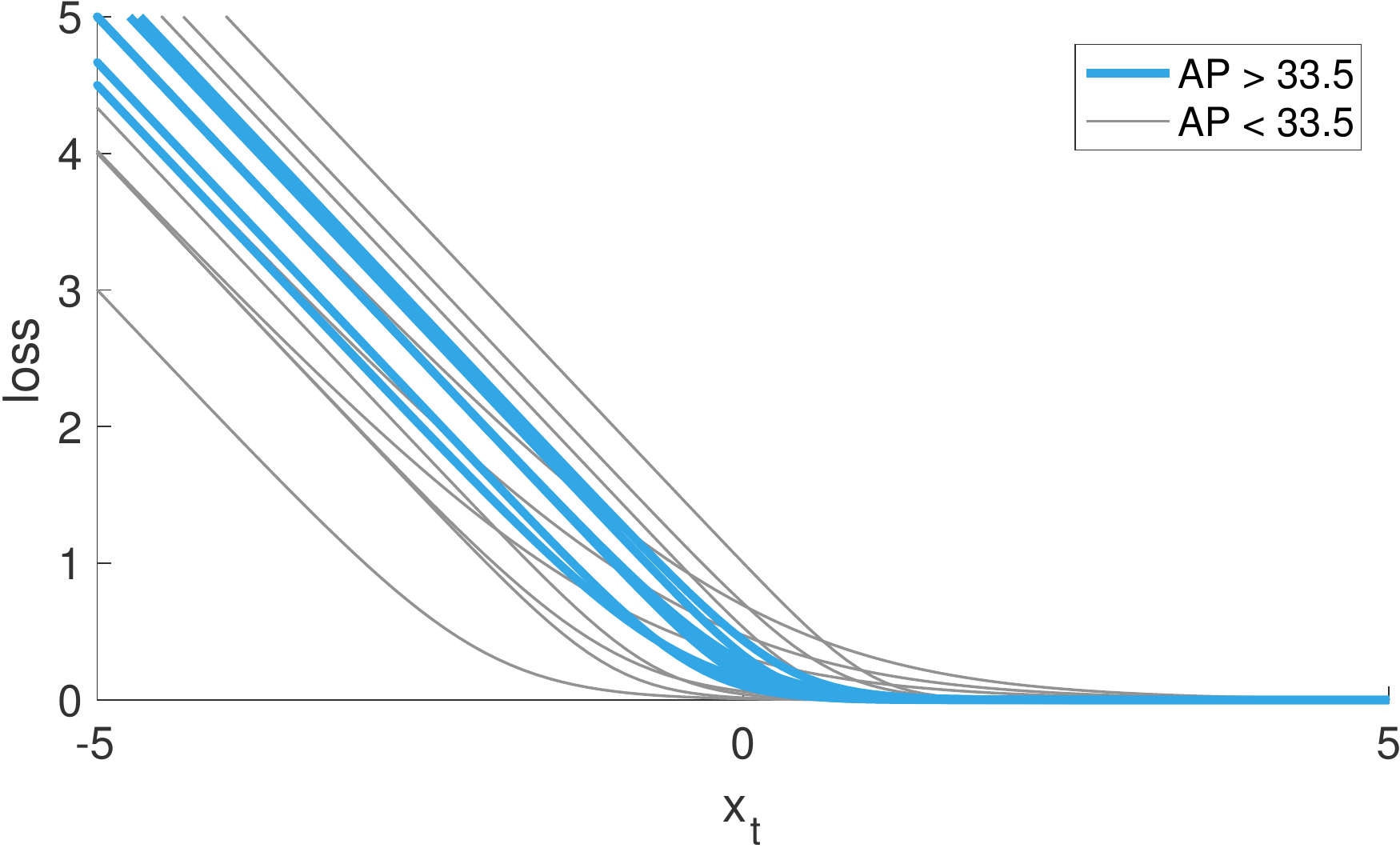}
\caption{Effectiveness of $\FQ$ with various settings $\gamma$ and $\beta$. The plots are color coded such that effective settings are shown in blue. }
\label{fig:loss-sweep}
\end{figure}

We found that various $\gamma$ and $\beta$ settings gave good results. In Figure~\ref{fig:loss-sweep} we show results for RetinaNet-50-600 with $\FQ$ for a wide set of parameters. The loss plots are color coded such that effective settings (models converged and with AP over 33.5) are shown in blue. We used $\alpha=.25$ in all experiments for simplicity. As can be seen, losses that reduce weights of well-classified examples ($\xt>0$) are effective.

More generally, we expect any loss function with similar properties as FL or $\FQ$ to be equally effective.

\section*{Appendix B: Derivatives}

For reference, derivates for CE, FL, and $\FQ$ \wrt $x$ are:
\begin{gather}
 \frac{d\CE}{dx} = y(\pt-1) \\
 \frac{d\FL}{dx} = y(1-\pt)^\gamma( \gamma \pt \log(\pt) + \pt - 1)\\
 \frac{d\FQ}{dx} = y(\pt^* -1)
\end{gather}
Plots for selected settings are shown in Figure~\ref{fig:loss-derivatives}. For all loss functions, the derivative tends to -1 or 0 for high-confidence predictions. However, unlike CE, for effective settings of both FL and $\FQ$, the derivative is small as soon as $\xt>0$.

{\small\bibliographystyle{ieee}\bibliography{dense.bib}}

\end{document}